\documentclass[a4paper,orivec,runningheads]{llncs}
\usepackage{makeidx}  
\usepackage{graphicx}
\usepackage{subfigure}
\usepackage{amsmath}
\usepackage{amssymb}
\usepackage{verbatim}
\usepackage{algorithm}
\usepackage{algorithmic}
\usepackage{booktabs}
\usepackage{multirow}
\usepackage{pbox}
\usepackage{dirtytalk}
\usepackage{hyperref}

\begin{document}

\pagenumbering{arabic}

%

%
%
\title{Automatic Brain Tumor Segmentation using Cascaded Anisotropic Convolutional Neural Networks}

\titlerunning{Automatic Brain Tumor Segmentation using Cascaded Anisotropic CNNs}  
%
\author{Guotai Wang \and Wenqi Li \and  S\'ebastien Ourselin \and Tom Vercauteren}
\institute{
Translational Imaging Group, CMIC, University College London, UK\\
Wellcome/EPSRC Centre for Interventional and Surgical Sciences,
UCL, London, UK
\email{guotai.wang.14@ucl.ac.uk}
}

\maketitle              

\begin{abstract}
A cascade of fully convolutional neural networks is proposed to segment multi-modal Magnetic Resonance (MR) images with brain tumor into background and three hierarchical regions: whole tumor, tumor core and enhancing tumor core.  The cascade is designed to decompose the multi-class segmentation problem into a sequence of three binary segmentation problems according to the subregion hierarchy. The whole tumor is segmented in the first step and the bounding box of the result is used for the tumor core segmentation in the second step. The enhancing tumor core is then segmented based on the bounding box of the tumor core segmentation result. Our networks consist of multiple layers of anisotropic and dilated convolution filters, and they are combined with multi-view fusion to reduce false positives.  Residual connections and multi-scale predictions are employed in these networks to boost the segmentation performance. Experiments with BraTS 2017 validation set show that the proposed method achieved average Dice scores of 0.7859, 0.9050, 0.8378 for enhancing tumor core, whole tumor and tumor core, respectively. The corresponding values for BraTS 2017 testing set were 0.7831, 0.8739, and 0.7748, respectively. 
\end{abstract}

\section{Introduction}
Gliomas are the most common brain tumors that arise from glial cells. They can be categorized into two basic grades: low-grade gliomas (LGG) that tend to exhibit benign tendencies and indicate a better prognosis for the patient, and high-grade gliomas (HGG) that are malignant and more aggressive. With the development of medical imaging, brain tumors can be imaged by various Magnetic Resonance (MR) sequences, such as T1-weighted, contrast enhanced T1-weighted (T1c), T2-weighted and Fluid Attenuation Inversion Recovery (FLAIR) images. Different sequences can provide complementary information to analyze different subregions of gliomas. For example, T2 and FLAIR highlight the tumor with peritumoral edema, designated \say{whole tumor} as per~\cite{Menze2015}. T1 and T1c highlight the tumor without peritumoral edema, designated \say{tumor core} as per~\cite{Menze2015}. An enhancing region of the tumor core with hyper-intensity can also be observed in T1c, designated \say{enhancing tumor core} as per~\cite{Menze2015}. 

Automatic segmentation of brain tumors and substructures has a potential to provide accurate and reproducible measurements of the tumors. It has a great potential for better diagnosis, surgical planning and treatment assessment of brain tumors~\cite{Menze2015,Bakas2017}. However, this segmentation task is challenging because 1) the size, shape, and localization of brain tumors have considerable variations among patients. This limits the usability and usefulness of prior information about shape and location that are widely used for robust segmentation of many other anatomical structures~\cite{Wang2015c,Grosgeorge2013}; 2) the boundaries between adjacent structures are often ambiguous due to the smooth intensity gradients, partial volume effects and bias field artifacts.

There have been many studies on automatic brain tumor segmentation over the past decades~\cite{Wang2014b}. Most current methods use generative or discriminative approaches. Generative approaches explicitly model the probabilistic distributions of anatomy and appearance of the tumor or healthy tissues~\cite{Kaus2001,Menze2010a}.  They often present good generalization to unseen images by incorporating domain-specific prior knowledge. However, accurate probabilistic distributions of brain tumors are hard to model. Discriminative approaches directly learn the relationship between image intensities and tissue classes, and they require a set of annotated training images for learning. Representative works include classification based on support vector machines~\cite{Lee2005} and decision trees~\cite{Zikic2012}. 

In recent years, discriminative methods based on deep neural networks have achieved
state-of-the-art performance for multi-modal brain tumor segmentation. In~\cite{Havaei2016}, a convolutional neural network (CNN) was proposed to exploit both local and more global features for robust brain tumor segmentation. However, their approach works on individual 2D slices without considering 3D contextual information.  DeepMedic~\cite{Kamnitsas2017} uses a dual pathway 3D CNN with 11 layers for brain tumor segmentation. The network processes the input image at multiple scales and the result is post-processed by a fully connected Conditional Random Field (CRF) to remove false positives. However, DeepMedic works on local image patches and has a low inference efficiency. Recently, several ideas to improve the segmentation performance of CNNs have been explored
in the literature. 3D U-Net~\cite{Abdulkadir2016} allows end-to-end training and testing for volumetric image segmentation. HighRes3DNet~\cite{Li2017} proposes a compact end-to-end 3D CNN structure that maintains high-resolution multi-scale features with dilated convolution and residual connection~\cite{Wang2017pami,Chen2016a}. Other works also propose using fully convolutional networks~\cite{B2016,Fidon2017a}, incorporating large visual contexts by employing a mixture of convolution and downsampling operations~\cite{Kamnitsas2017,Havaei2016}, and handling imbalanced training data by designing new loss functions~\cite{Fidon2017b,Sudre2017} and sampling strategies~\cite{Sudre2017}.

The contributions of this work are three-fold. First, we propose a cascade of CNNs to segment brain tumor
subregions sequentially. The cascaded CNNs separate the complex problem of multiple class segmentation into  three simpler binary segmentation problems, and take advantage of the hierarchical structure of tumor subregions to reduce false positives. Second, we propose a novel network structure with anisotropic convolution to deal with 3D images as a trade-off among receptive field, model complexity and memory consumption. 
It uses dilated convolution, residual connection and multi-scale prediction to improve segmentation performance. Third, we propose to fuse the output of CNNs in three orthogonal views for more robust segmentation of brain tumor. 
\section{Methods}
\begin{figure}[t]
	\centering
	\includegraphics[width=1.0\linewidth]{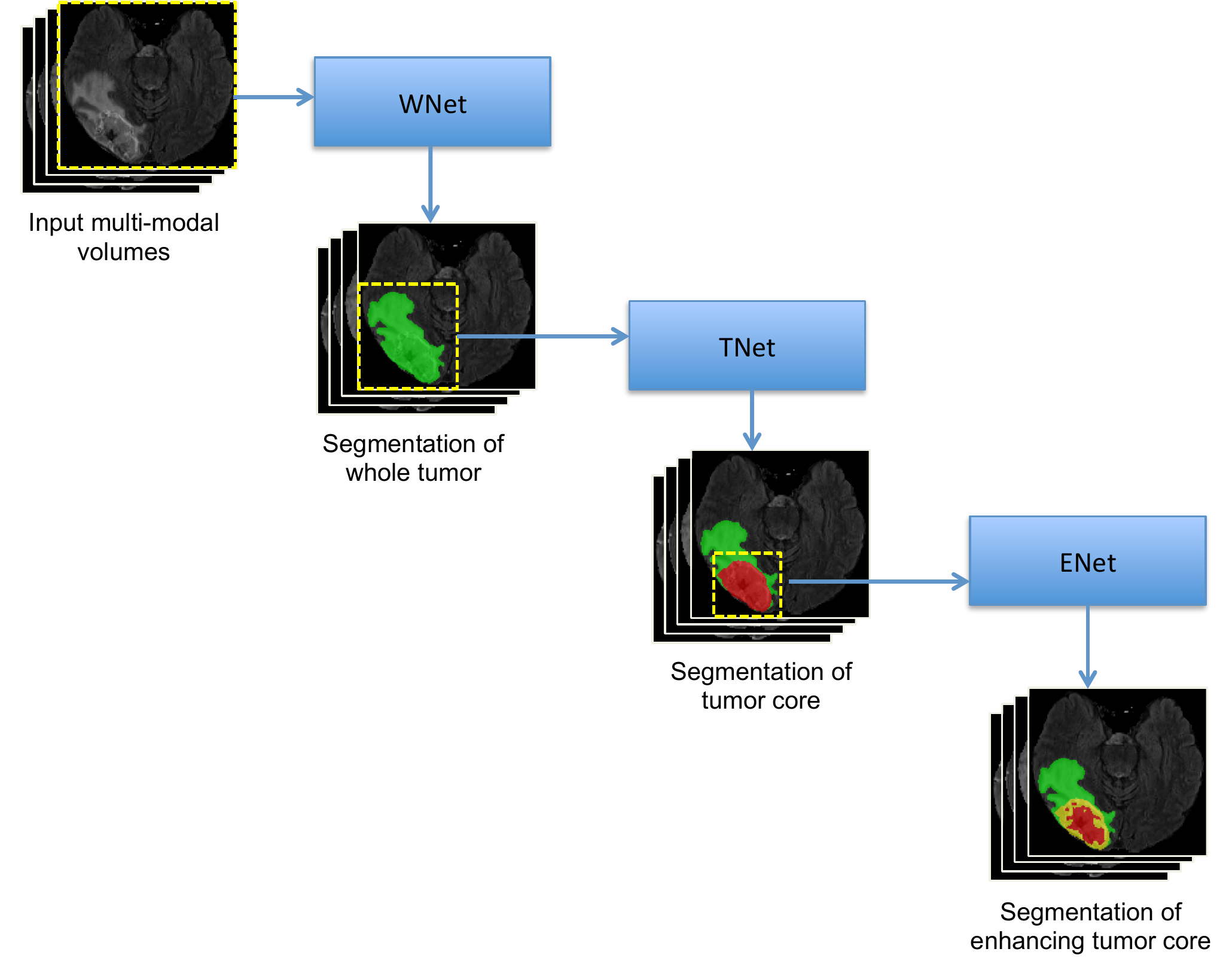}
	\caption[The proposed segmentation method]{
		The proposed triple cascaded framework for brain tumor segmentation. Three networks are proposed to hierarchically segment whole tumor (WNet), tumor core (TNet) and enhancing tumor core (ENet) sequentially.
	}
	\label{fig:cascaded_framework}
\end{figure}
\subsection{Triple Cascaded Framework}
The proposed cascaded framework is shown in Fig.~\ref{fig:cascaded_framework}. We use three networks to hierarchically and sequentially segment substructures of brain tumor, and each of these networks deals with a binary segmentation problem. The first network (WNet) segments the whole tumor from multi-modal 3D volumes of the same patient. Then a bounding box of the whole tumor is obtained. The cropped region of the input images based on the bounding box is used as the input of the second network (TNet) to segment the tumor core. Similarly, image region inside the bounding box of the tumor core is used as the input of the third network (ENet) to segment the enhancing tumor core. In the training stage, the bounding boxes are automatically generated based on the ground truth. In the testing stage, the bounding boxes are generated based on the binary segmentation results of the whole tumor and the tumor core, respectively. The segmentation result of WNet is used as a crisp binary mask for the output of TNet, and the segmentation result of TNet is used as a crisp binary mask for the output of ENet, which serves as anatomical constraints for the segmentation.

\subsection{Anisotropic Convolutional Neural Networks}
\begin{figure}[t]
	\centering
	\includegraphics[width=1.0\linewidth]{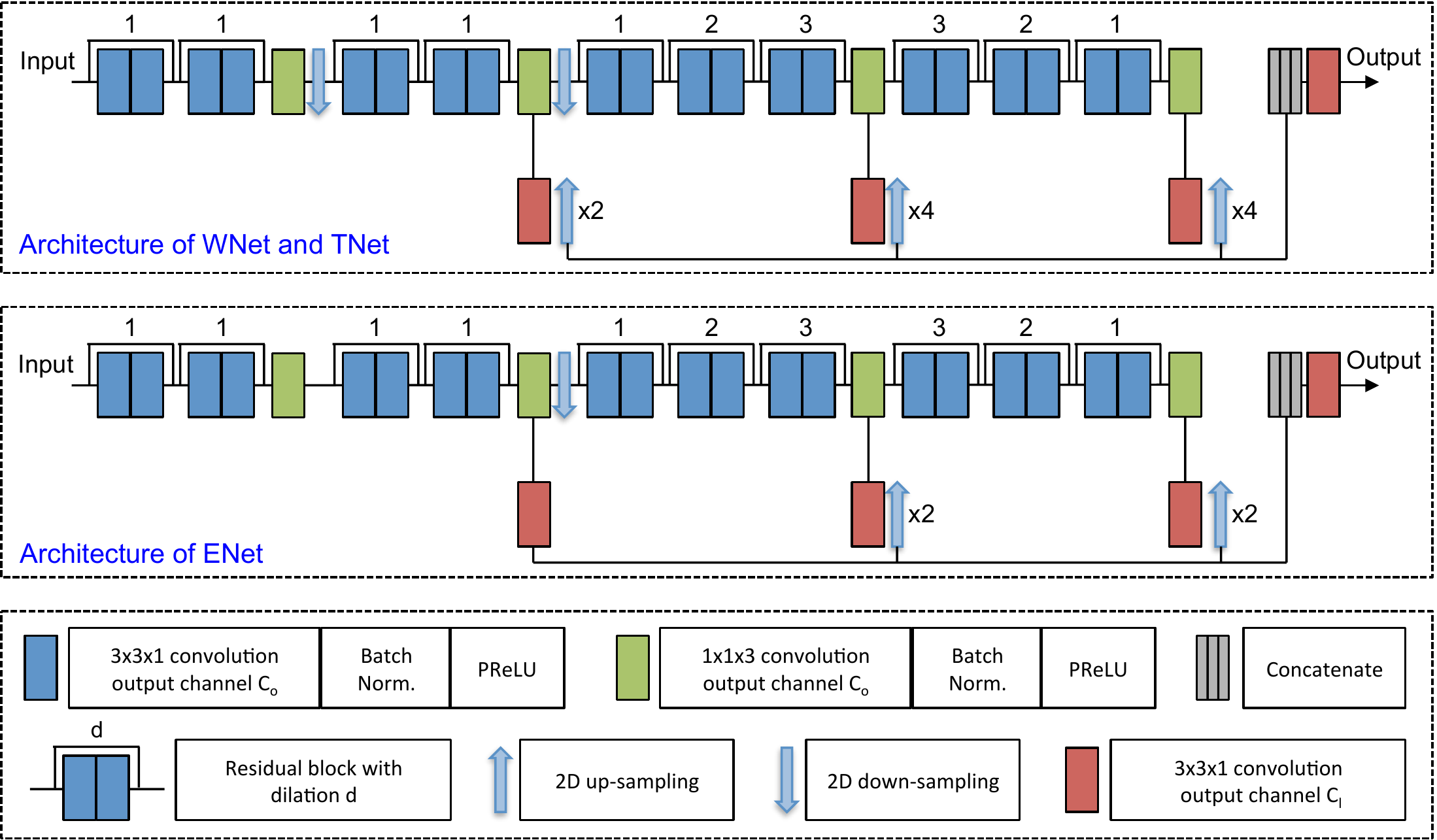}
	\caption[anisotropic network]{
		Our anisotropic convolutional networks with dilated convolution, residual connection and multi-scale fusion. ENet uses only one downsampling layer considering its smaller input size.
	}
	\label{fig:network}
\end{figure}
For 3D neural networks, the balance among receptive field, model complexity and memory consumption should be considered. A small receptive field leads the model to only use local features, and a larger receptive field allows the model to also learn more global features. Many 2D networks use a very large receptive field to capture features from the entire image context, such as FCN~\cite{Long2014} and U-Net~\cite{Hefny2015a}. They require a large patch size for training and testing. Using a large 3D receptive field also helps to obtain more global features for 3D volumes. However, the resulting large 3D patches for training consume a lot of memory, and therefore restrict the resolution and number of features in the network, leading to limited model complexity and low representation ability. As a trade-off, we propose anisotropic networks that take a stack of slices as input with a large receptive field in 2D and a relatively small receptive field in the out-plane direction that is orthogonal to the 2D slices. The 2D receptive fields for WNet, TNet and ENet are 217$\times$217, 217$\times$217, and 113$\times$113, respectively. During training and testing, the 2D sizes of the inputs are typically smaller than the corresponding 2D receptive fields. WNet, TNet and ENet have the same out-plane receptive field of 9. The architectures of these proposed networks are shown in Fig.~\ref{fig:network}. All of them are fully convolutional and use 10 residual connection blocks with anisotropic convolution, dilated convolution, and multi-scale prediction.

\subsubsection{Anisotropic and Dilated Convolution.}
To deal with anisotropic receptive fields, we decompose a 3D kernel with a size of 3$\times$3$\times$3 into an intra-slice kernel with a size of 3$\times$3$\times$1 and an inter-slice kernel with a size of  1$\times$1$\times$3. Convolution layers with either of these kernels have $C_o$ output channels and each is followed by a batch normalization layer and an activation layer, as illustrated by blue and green blocks in Fig.~\ref{fig:network}. The activation layers use Parametric Rectified Linear Units (PReLU) that have been shown better performance than traditional rectified units~\cite{He2015iccv} .  WNet and TNet use 20 intra-slice convolution layers and four inter-slice convolution layers with two 2D downsampling layers. ENet uses the same set of convolution layers as WNet but only one downsampling layer considering its smaller input size. We only employ up to two layers of downsampling in order to avoid large image resolution reduction and loss of segmentation details. After the downsampling layers, we use dilated convolution for intra-slice kernels to enlarge the receptive field within a slice. The dilation parameter is set to 1 to 3 as shown in Fig.~\ref{fig:network}. 

\subsubsection{Residual Connection.} For effective training of deep CNNs, residual connections~\cite{He2015res} were introduced to create identity mapping connections to bypass the parameterized layers in a network. Our WNet, TNet and ENet have 10 residual blocks. Each of these blocks contains two intra-slice convolution layers, and the input of a residual block is directly added to the output, encouraging the block to learn residual functions with reference to the input. This can make information propagation smooth and speed the convergence of training~\cite{He2015res,Li2017}.

\subsubsection{Multi-scale Prediction.}  With the kernel sizes used in our networks, shallow layers learn to represent local and low-level features while deep layers learn to represent more global and high-level features. To combine features at different scales, we use three 3$\times$3$\times$1 convolution layers at different depths of the network to get multiple intermediate predictions and upsample them to the resolution of the input. A concatenation of these predictions are fed into an additional 3$\times$3$\times$1 convolution layer to obtain the final score map. These layers are illustrated by red blocks in Fig.~\ref{fig:network}. The outputs of these layers have $C_l$ channels where $C_l$ is the number of classes for segmentation in each network. $C_l$ equals to 2 in our method. A combination of predictions from multiple scales has also been used in~\cite{Xie2015,Fidon2017b}. 

\subsubsection{Multi-view Fusion.}
\begin{figure}[t]
	\centering
	\includegraphics[width=0.8\linewidth]{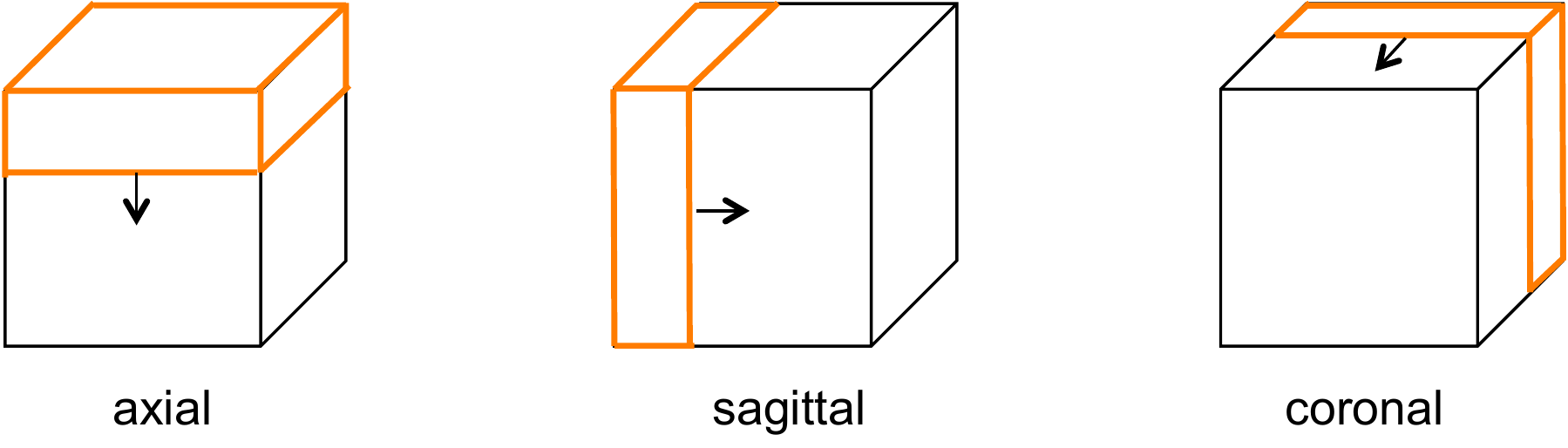}
	\caption[anisotropic network]{
		Illustration of multi-view fusion at one level of the proposed cascade. Due to the anisotropic receptive field of our networks, we average the softmax outputs in axial, sagittal and coronal views. The orange boxes show examples of sliding windows for testing. Multi-view fusion is implemented for WNet, TNet, and ENet, respectively.
	}
	\label{fig:fusion}
\end{figure}

Since the anisotropic convolution has a small receptive field in the out-plane direction, to take advantage of 3D contextual information, we fuse the segmentation results from three different orthogonal views. Each of WNet, TNet and ENet was trained in axial, sagittal and coronal views respectively. During the testing time, predictions in these three views are fused to get the final segmentation. For the fusion, we average the softmax outputs in these three views for each level of the cascade of WNet, TNet, and ENet, respectively. An illustration of multi-view fusion at one level is shown in Fig.~\ref{fig:fusion}.

\begin{figure}[t]
	\centering
	\includegraphics[width=1.0\linewidth]{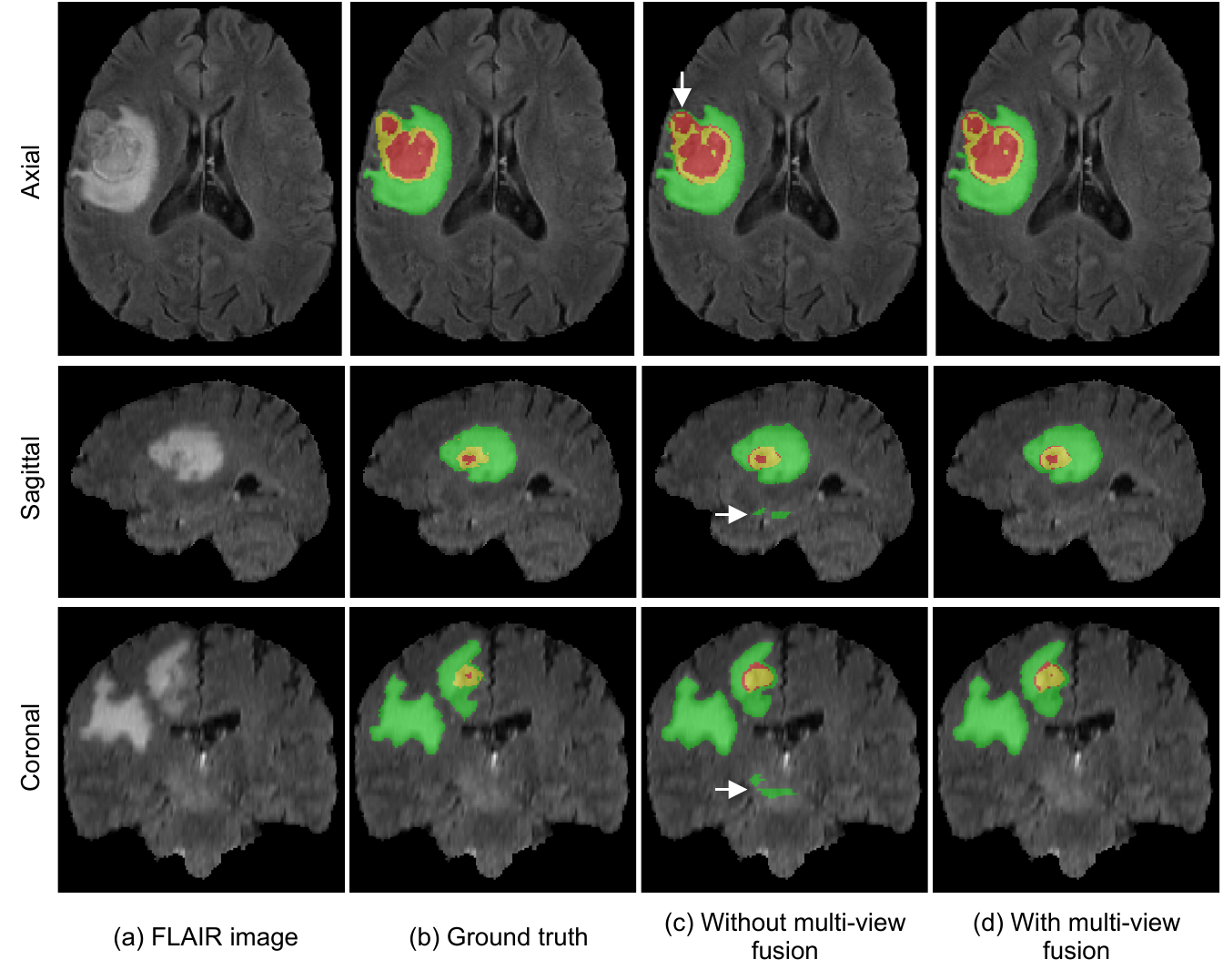}
	\caption[Segmentation result of the brain tumor (HGG) from a training image.]{
		Segmentation result of the brain tumor (HGG) from a training image. Green: edema; Red: non-enhancing tumor core; Yellow: enhancing tumor core. (c) shows the result when only networks in axial view are used, with mis-segmentations highlighted by white arrows. (d) shows the result with multi-view fusion.
	}
	\label{fig:hgg}
\end{figure}
\begin{figure}[t]
	\centering
	\includegraphics[width=1.0\linewidth]{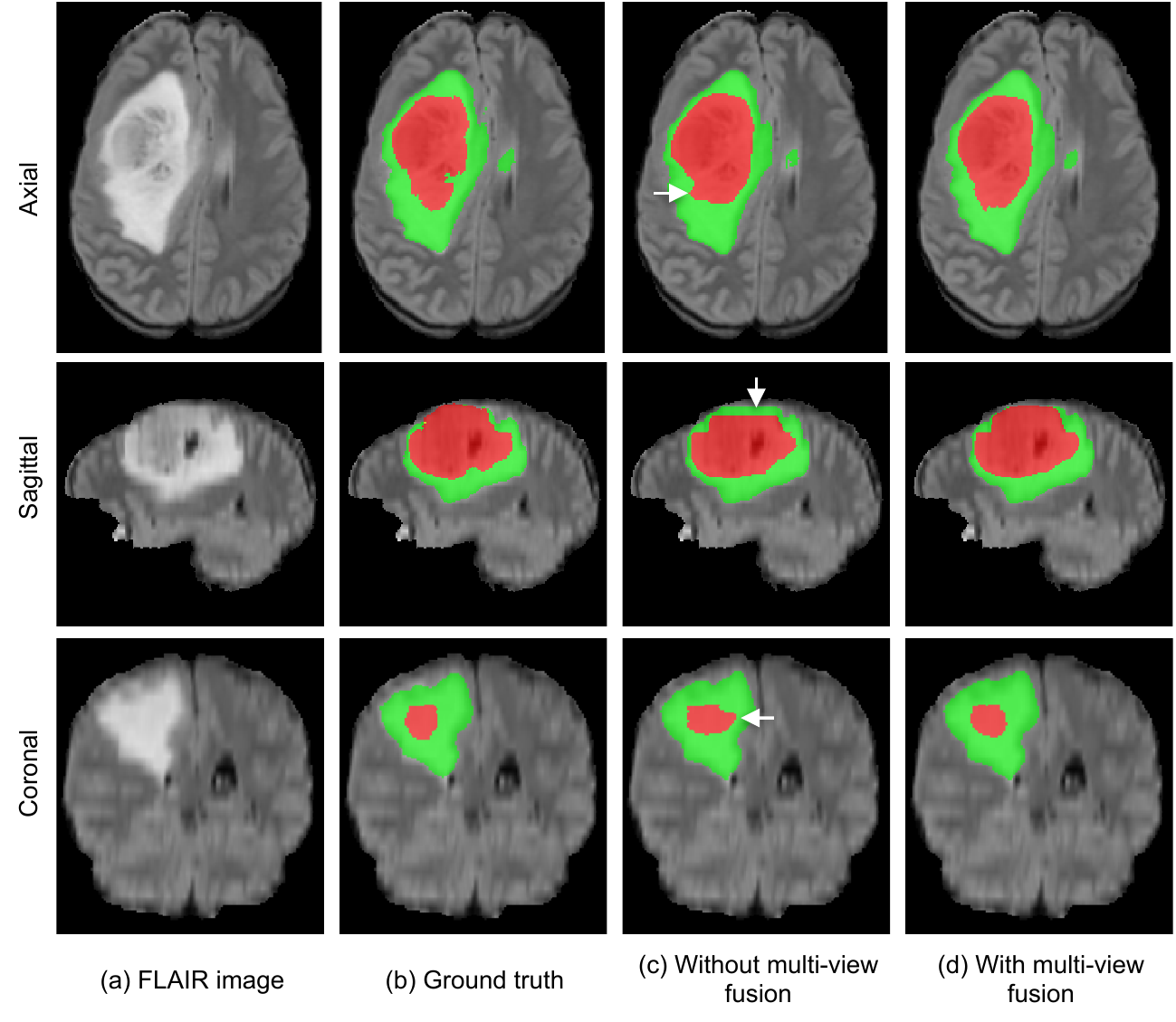}
	\caption[Segmentation result of the brain tumor (LGG) from a training image]{
		Segmentation result of the brain tumor (LGG) from a training image. Green: edema; Red: non-enhancing tumor core; Yellow: enhancing tumor core. (c) shows the result when only networks in axial view are used, with mis-segmentations highlighted by white arrows. (d) shows the result with multi-view fusion.
	}
	\label{fig:lgg}
\end{figure}

\section{Experiments and Results}

\subsubsection{Data and Implementation Details.}

We used the BraTS 2017\footnote{\url{http://www.med.upenn.edu/sbia/brats2017.html}}~\cite{Menze2015,Bakas2017,Bakas2017a,Bakas2017b} dataset for experiments. The training set contains images from 285 patients (210 HGG and 75 LGG). The BraTS 2017 validation and testing set contain images from 46 and 146 patients with brain tumors of unknown grade, respectively. Each patient was scanned with four sequences: T1, T1c, T2 and FLAIR. All the images were skull-striped and re-sampled to an isotropic 1mm$^3$ resolution, and the four sequences of the same patient had been co-registered. The ground truth were obtained by manual segmentation results given by experts. We uploaded the segmentation results obtained by the experimental algorithms to the BraTS 2017 server, and the server provided quantitative evaluations including Dice score and Hausdorff distance compared with the ground truth.

Our networks were implemented in Tensorflow\footnote{\url{https://www.tensorflow.org}}~\cite{Abadi2016} using NiftyNet\footnote{\url{http://niftynet.io}}\footnote{\url{https://cmiclab.cs.ucl.ac.uk/CMIC/NiftyNet/tree/dev/demos/BRATS17}}~\cite{Gibson2017a}. We used Adaptive Moment Estimation (Adam)~\cite{Kingma2015} for training, with initial learning rate $10^{-3}$, weight decay $10^{-7}$, batch size 5, and maximal iteration 30k. Training was implemented on
an NVIDIA TITAN X GPU. The training patch size was 144$\times$144$\times$19, 96$\times$96$\times$19, and 64$\times$64$\times$19 for WNet, TNet and ENet, respectively. We set $C_o$ to 32 and $C_l$ to 2 for these three types of networks.  For pre-processing, the images were normalized by the mean value and standard deviation of the training images for each sequence. We used the Dice loss function~\cite{Milletari2016,Fidon2017b} for training of each network.
 
\subsubsection{Segmentation Results.} 
Fig.~\ref{fig:hgg} and Fig.~\ref{fig:lgg} show examples for HGG and LGG segmentation from training images, respectively. In both figures, for simplicity of visualization, only the FLAIR image is shown. The green, red and yellow colors show the edema, non-enhancing and enhancing tumor cores, respectively. We compared the proposed method with its variant that does not use multi-view fusion. For this variant, we trained and tested the networks only in axial view. In Fig.~\ref{fig:hgg}, segmentation without and with multi-view fusion are presented in the third and forth columns, respectively.  It can be observed that the segmentation without multi-view fusion shown in Fig.~\ref{fig:hgg}(c) has some noises for edema and enhancing tumor core, which is highlighted by white arrows. In contrast, the segmentation with multi-view fusion shown in Fig.~\ref{fig:hgg}(d) is more accurate. In Fig.~\ref{fig:lgg}, the LGG does not contain enhancing regions. The two counterparts achieve similar results for the whole tumor. However, it can be observed that the segmentation of tumor core is more accurate by using multi-view fusion.

\begin{table}
	\centering
	\small
	\caption{Mean values of Dice and Hausdorff measurements of the proposed method on BraTS 2017 validation set. EN, WT, TC denote enhancing tumor core, whole tumor and tumor core, respectively.}
	\label{tab:valid}
	\begin{tabular}{l|c|c|c|c|c|c}
		\hline
		& \multicolumn{3}{c|}{Dice} & \multicolumn{3}{c}{Hausdorff (mm)}  \\ \hline
		& ET & WT & TC & ET & WT & TC \\ \hline
		Without multi-view fusion & 0.7411 & 0.8896 & 0.8255 & 5.3178 & 12.4566 & 9.6616 \\
		With multi-view fusion & 0.7859 & 0.9050 & 0.8378 & 3.2821 & 3.8901 & 6.4790 \\
		\hline
	\end{tabular}
\end{table}
\begin{table}
	\centering
	\small
	\caption{Dice and Hausdorff measurements of the proposed method on BraTS 2017 testing set. EN, WT, TC denote enhancing tumor core, whole tumor and tumor core, respectively.}
	\label{tab:test}
	\begin{tabular}{l|c|c|c|c|c|c}
		\hline
		& \multicolumn{3}{c|}{Dice} & \multicolumn{3}{c}{Hausdorff (mm)}  \\ \hline
		& ET & WT & TC & ET & WT & TC \\ \hline
		Mean & 0.7831 & 0.8739 & 0.7748 & 15.9003 & 6.5528 & 27.0472 \\
		Standard deviation & 0.2215 & 0.1319 & 0.2700 & 67.8552 & 10.6915 & 84.4297 \\
		Median & 0.8442 & 0.9162 & 0.8869 & 1.7321 & 3.0811 & 3.7417\\
25 quantile & 0.7287 & 0.8709 & 0.7712 & 1.4142 & 2.2361 & 2.0000 \\	
75 quantile & 0.8882 & 0.9420 & 0.9342 & 3.1217 & 5.8310 & 8.4255 \\		\hline
	\end{tabular}
\end{table}

Table~\ref{tab:valid} presents quantitative evaluations with the BraTS 2017 validation set. It shows that the proposed method achieves average Dice scores of 0.7859, 0.9050 and 0.8378 for enhancing tumor core, whole tumor and tumor core, respectively. For comparison, the variant without multi-view fusion obtains average Dice scores of 0.7411, 0.8896 and 0.8255 fore these three regions, respectively. 

Table~\ref{tab:test} presents quantitative evaluations with the BraTS 2017 testing set. It shows the mean values, standard deviations, medians, 25 and 75 quantiles of Dice and Hausdorff distance. Compared with the performance on the validation set, the performance on the testing set is lower, with average Dice scores of 0.7831, 0.8739, and 0.7748 for enhancing tumor core, whole tumor and tumor core, respectively. The higher median values show that good segmentation results are achieved for most images, and some outliers contributed to the lower average scores. 


\section{Discussion and Conclusion}
There are several benefits of using a cascaded framework for segmentation of hierarchical structures~\cite{Christ2016}. First, compared with using a single network for all substructures of the brain tumor that requires complex network architectures, using three binary segmentation networks allows for a simpler network for each task. Therefore, they are easier to train and can reduce over-fitting. Second, the cascade helps to reduce false positives since TNet works on the region extracted by WNet, and ENet works on the region extracted by TNet. Third, the hierarchical pipeline follows anatomical structures of the brain tumor and uses them as spatial constraints. The binary crisp masks restrict the tumor core to be inside the whole tumor region and enhancing tumor core to be inside the tumor core region, respectively.
In~\cite{Fidon2017b}, the hierarchical structural information was leveraged to design a Generalised Wasserstein Dice loss function for imbalanced multi-class segmentation. However, that work did not use the hierarchical structural information as spatial constraints. One drawback of the cascade is that it is not end-to-end and requires longer time for training and testing compared with its multi-class counterparts using similar structures. However, we believe this is not an important issue for automatic brain tumor segmentation. Also, at inference time, our framework is more computationally efficient than most competitive approaches including ScaleNet~\cite{Fidon2017a} and  DeepMedic~\cite{Kamnitsas2017}. 

The results show that our proposed method achieved competitive performance for automatic brain tumor segmentation. Fig.~\ref{fig:hgg} and Fig.~\ref{fig:lgg} demonstrate that the multi-view fusion helps to improve segmentation accuracy. This is mainly because our networks are designed with anisotropic receptive fields. The multi-view fusion is an ensemble of networks in three orthogonal views, which takes advantage of 3D contextual information to obtain higher accuracy. Considering the different imaging resolution in different views, it may be more reasonable to use a weighted average of axial, sagittal and coronal views rather than a simple average of them in the testing stage~\cite{Mortazi2017}. Our current results are not post-processed by CRFs that have been shown effective to get more spatially regularized segmentation~\cite{Kamnitsas2017}. Therefore, it is of interest to further improve our segmentation results by using CRFs.  

In conclusion, we developed a cascaded system to segment glioma subregions from multi-modal
brain MR images. We convert the multi-class segmentation problem to three cascaded binary segmentation problems, and use three networks to segment the whole tumor, tumor core and enhancing tumor core, respectively. Our networks use an anisotropic structure, which considers the balance among receptive field, model complexity and memory consumption. We also use multi-view fusion to reduce noises in the segmentation result.  Experimental results on BraTS 2017 validation set show that the proposed method achieved average Dice scores of 0.7859, 0.9050, 0.8378 for enhancing tumor core, whole tumor and tumor core, respectively. The corresponding values for BraTS 2017 testing set were 0.7831, 0.8739, and 0.7748, respectively. 

\subsubsection{Acknowledgements.}\label{sec:acknowledgements}
We would like to thank the NiftyNet team. This work was supported through an Innovative Engineering for Health award by the Wellcome Trust [WT101957], Engineering and Physical Sciences Research Council (EPSRC) [NS/A000027/1], the National Institute for Health Research University College London Hospitals Biomedical Research Centre (NIHR BRC UCLH/UCL High Impact Initiative), a UCL Overseas Research Scholarship, a UCL Graduate Research Scholarship, hardware donated by NVIDIA, and the Health Innovation Challenge Fund [HICF-T4-275, WT 97914], a parallel funding partnership between the Department of Health and Wellcome Trust.
%
%
\bibliographystyle{splncs03}
\bibliography{reference/miccai2017}


%
%

\end{document}